\documentclass{article}

\usepackage{times}
\usepackage{graphicx} 
\usepackage{subfigure} 

\usepackage{natbib}

\usepackage{algorithm}
\usepackage{algorithmic}

\usepackage{ amssymb }
\usepackage{amsmath}
\usepackage{booktabs}       
\usepackage{float}
\usepackage{todonotes}

\usepackage{hyperref}

\usepackage[accepted]{icml2017}

\icmltitlerunning{Meta Networks}

\begin{document} 

\twocolumn[
\icmltitle{Meta Networks}



\icmlsetsymbol{equal}{*}

\begin{icmlauthorlist}
\icmlauthor{Tsendsuren Munkhdalai}{to}
\icmlauthor{Hong Yu}{to}
\end{icmlauthorlist}

\icmlaffiliation{to}{University of Massachusetts, MA, USA}

\icmlcorrespondingauthor{Tsendsuren Munkhdalai}{tsendsuren.munkhdalai@umassmed.edu}

\icmlkeywords{Meta learning, learning to learn, one-shot learning, deep neural networks, fast transfer learning}

\vskip 0.3in
]



\printAffiliationsAndNotice{}  

\begin{abstract} 
Neural networks have been successfully applied in applications with a large amount of labeled data. However, the task of rapid generalization on new concepts with small training data while preserving performances on previously learned ones still presents a significant challenge to neural network models. In this work, we introduce a novel meta learning method, Meta Networks (MetaNet), that learns a meta-level knowledge across tasks and shifts its inductive biases via fast parameterization for rapid generalization.
When evaluated on Omniglot and Mini-ImageNet benchmarks, our MetaNet models achieve a near human-level performance and outperform the baseline approaches by up to 6\% accuracy. We demonstrate several appealing properties of MetaNet relating to generalization and continual learning.
\end{abstract} 

\section{Introduction}
\label{introduction}

Deep neural networks have shown great success in several application domains when a large amount of labeled data is available for training. 
However, the availability of such large training data has generally been a prerequisite in a majority of learning tasks. 
Furthermore, the standard deep neural networks lack the ability to continuous learning or incrementally learning new concepts on the fly, without forgetting or corrupting previously learned patterns. In contrast, humans can rapidly learn and generalize from a few examples of the same concept. Humans are also very good at incremental (i.e. continuous) learning. These abilities have been mostly explained by the meta learning (i.e. learning to learn) process in the brain \cite{harlow1949formation}.

\begin{figure}[t]
\vskip 0.2in
\begin{center}
\centerline{\includegraphics[width=1.0\columnwidth]{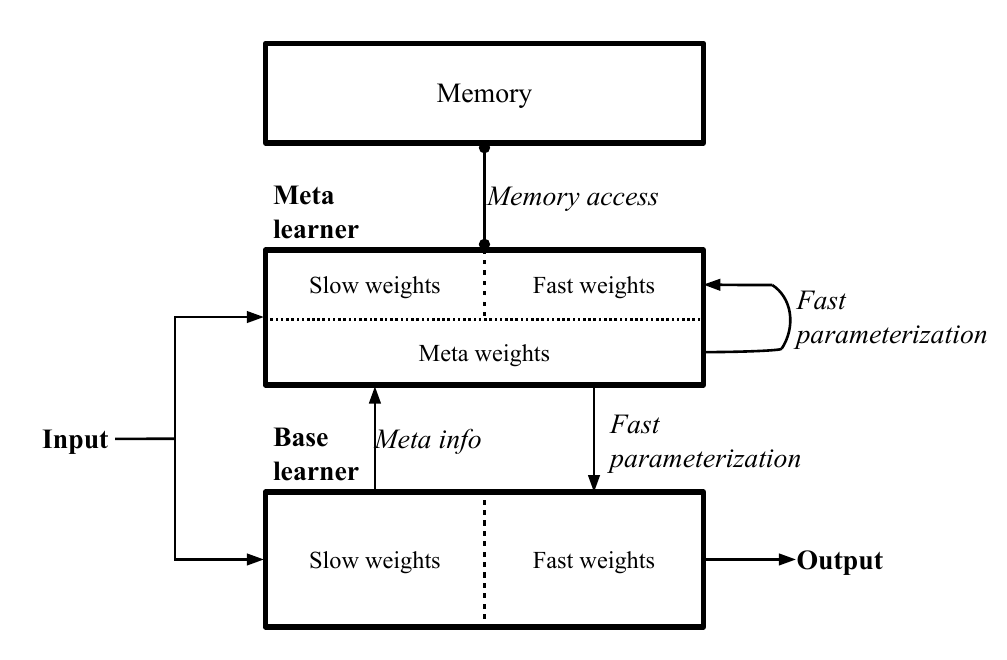}}
\caption{Overall architecture of Meta Networks.}
\label{metanet}
\end{center}
\vskip -0.2in
\end{figure} 

Previous work on meta learning has formulated the problem as two-level learning, a slow learning of a meta-level model performing across tasks and a rapid learning of a base-level model acting within each task \cite{mitchell1993explanation,vilalta2002perspective}. The goal of a meta-level learner is to acquire generic knowledge of different tasks. The knowledge can then be transferred to the base-level learner to provide generalization in the context of a single task. The base and meta-level models can be framed in a single learner \citep{schmidhuber1987} or in separate learners \citep{bengio1990learning,hochreiter2001learning}.

In this work we introduce a meta learning model called MetaNet (for Meta Networks) that supports meta-level continual learning by allowing neural networks to learn and to generalize a new task or concept from a single example on the fly. The overall architecture of MetaNet is shown in Figure \ref{metanet}. MetaNet consists of two main learning components, a base learner and a meta learner, and is equipped with an external memory. Learning occurs at two levels in separate spaces (i.e. meta space and task space). The base learner performs in the input task space whereas the meta learner operates in a task-agnostic meta space. By operating in the abstract meta space, the meta learner supports continual learning and performs meta knowledge acquisition across different tasks. Towards this end, the base learner first analyzes the input task. The base learner then provides the meta learner with a feedback in the form of higher order meta information to explain its own status in the current task space. Based on the meta information, the meta learner rapidly parameterizes both itself and the base learner so that the MetaNet model can recognize the new concepts of the input task. Specifically, the training weights of MetaNet evolve at different time-scales: standard slow weights are updated through a learning algorithm (i.e. REINFORCE),  task-level fast weights are updated within the scope of each task, and example-level fast weights are updated for a specific input example. Finally MetaNet equipped with external memory allows for rapid learning and generalization.

Under the MetaNet framework, it is important to define the types of the meta information which can be obtained from the learners. While other representations of meta information are also applicable, we use loss gradients as meta information. MetaNet has two types of loss functions with distinct objectives: a representation (i.e. embedding) loss defined for the good representation learner criteria and a main (task) loss used for the input task objective.

We extensively studied the performance and the characteristics of MetaNet on one-shot supervised learning (SL) problems under several different settings. Our proposed method not only improves the state-of-the-art results on the standard benchmarks, but also shows some interesting properties related to generalization and continual learning.

\section{Related Work}
\label{related work}

Our work connects different threads of research in order to model neural architectures for 
rapid learning and generalization. Rapid learning and generalization refers to a one-shot learning scenario where a learner is introduced to a sequence of tasks, where each task entails multi-class classification with a single or few labeled example per class. A key challenge in this setting is that the classes or concepts vary across the tasks. Due to this, one-shot learning problems have been widely addressed by generative models and metric learning methods. One notable success is reported by a probabilistic programming approach \citep{lake2015human}. They used specific knowledge of how pen strokes are composed to produce characters of different alphabets. \citet{koch2015siamese} applied Siamese Networks to perform one-shot classification. Recently, \citet{vinyals2016matching} unified the training and testing of a one-shot learner under the same procedure and developed an end-to-end differentiable nearest neighbor method for one-shot learning. \citet{santoro2016meta} proposed a memory-based approach and trained Neural Turing Machines \cite{graves2014neural} for one-shot learning, although the meta-learner and the one-shot learner in this work are not separable explicitly. The training procedure used by \citet{santoro2016meta} adapted the work of \citet{hochreiter2001learning} in which they use LSTMs as the meta-level model. More recently an LSTM-based one-shot optimizer was proposed \cite{Sachin2017}. By taking in the loss, the gradient and the parameters of the base learner, the meta optimizer was trained to update the parameters for one-shot classification.

A related line of work focuses on building meta optimizers \cite{hochreiter2001learning,maclaurin2015gradient,andrychowicz2016learning,ke2017}. As the main interest here is to train an optimization algorithm within the meta learning framework, these efforts have mainly focused on tasks with large datasets. In contrast, with the absence of large datasets, our experimental setup emphasizes the difficulties of optimizing a neural network with a large number of parameters to generalize with limited examples of a new concept. Our work proposes a novel rapid parameterization approach by employing meta information. By following the success of the previous work \cite{mitchell1993explanation,younger1999fixed,andrychowicz2016learning,Sachin2017}, we study the meta information present in the loss gradient of neural nets.
Fast weights and utilizing one neural network to generate parameters for another neural network have previously been studied separately.
\citet{hinton1987using} suggested the usage of fast weights for rapid learning. \citet{ba2016using} recently used fast weights to replace soft attention mechanism. Fast weights have also been used to implement recurrent nets \cite{schmidhuber1992learning,Schmidhuber1993} and self-referential networks \cite{schmidhuber1987,Schmidhuber1993b}. These usages of fast weights are well motivated by the fact that synapses have dynamics at many different time-scales \cite{greengard2001neurobiology}.

The approach proposed by \citet{gomez2005evolving} is more closely related to our work. They used recurrent nets to generate fast weights for a single-layer network controller. \citet{de2016dynamic} used one network to generate slow filter weights for a convolutional neural net. More recently \citet{david2017} generated slow weights for recurrent nets. Our MetaNet generates fast weights at two time-scales by operating in meta space. To integrate the fast weights with the slow weights, we propose a novel layer augmentation approach.

Finally, we note that our MetaNet equipped with an external memory can be seen as a memory augmented neural network (MANN). MANNs have shown promising results on a range of tasks starting from small programming problems \cite{graves2014neural} to large-scale language tasks \cite{weston:15,sukhbaatar2015end,munkhdalai2016neural}.

\begin{algorithm}[t]
\caption{MetaNet for one-shot supervised learning}
\label{alg:metanet}
\begin{algorithmic}[1]
{\footnotesize
\REQUIRE Support set $\lbrace x'_i,y'_i \rbrace^N_{i=1}$ and Training set $\lbrace x_i,y_i \rbrace^L_{i=1}$
\REQUIRE Base learner $b$, Dynamic representation learning function $u$, Fast weight generation functions $m$ and $d$, and Slow weights $\theta=\lbrace W,Q,Z,G \rbrace$ 
\REQUIRE Layer augmentation scheme
\STATE Sample $T$ examples from support set
\FOR{$i=1, T$}
    \STATE $\mathcal{L}_{i} \leftarrow loss_{emb}(u(Q,x'_i),y'_i)$
    \STATE $\nabla_{i} \leftarrow \nabla_Q \mathcal{L}_i$
\ENDFOR
\STATE $Q^* = d(G, \lbrace \nabla \rbrace^T_{i=1})$
\FOR{$i=1, N$}
    \STATE $\mathcal{L}_{i} \leftarrow loss_{task}(b(W,x'_i),y'_i)$
    \STATE $\nabla_{i} \leftarrow \nabla_W \mathcal{L}_i$
    \STATE $W_i^* \leftarrow m(Z,\nabla_{i})$
    \STATE Store $W_i^*$ in $i$\textsuperscript{th} position of memory $M$
    \STATE $r'_i = u(Q,Q^*,x'_i)$
    \STATE Store $r'_i$ in $i$\textsuperscript{th} position of index memory $R$
\ENDFOR
\STATE $\mathcal{L}_{train}=0$
\FOR{$i=1, L$}
    \STATE $r_i = u(Q,Q^*,x_i)$
	\STATE $a_i = attention(R,r_i)$
	\STATE $W_i^* = softmax(a_i)^{\top} M$
    \STATE $\mathcal{L}_{train} \leftarrow \mathcal{L}_{train} + loss_{task}(b(W,W_{i}^*,x_i), y_i)$ \COMMENT{Alternatively the base learner can take as input $r_i$ instead of $x_i$}
\ENDFOR
\STATE Update $\theta$ using $\nabla_{\theta} \mathcal{L}_{train}$
}
\end{algorithmic}
\end{algorithm}

\section{Meta Networks}
\label{meta networks}

MetaNet learns to fast parameterize underlying neural networks for rapid generalizations by processing a higher order meta information, resulting in a flexible AI model that can adapt to a sequence of tasks with possibly distinct input and output distributions. The model consists of two main learning modules (Figure \ref{metanet}). The meta learner is responsible for fast weight generation by operating across tasks while the base learner performs within each task by capturing the task objective. The generated fast weights are integrated into both base learner and meta learner to shift the inductive bias of the learners. We propose a novel layer augmentation method to integrate the standard slow weights and the task or example specific fast weights in a neural net.

To train MetaNet, we adapt a task formulation procedure by \citet{vinyals2016matching}. We form a sequence of tasks, where each task consists of a support set $\lbrace x'_i,y'_i \rbrace^N_{i=1}$ and a training set $\lbrace x_i,y_i \rbrace^L_{i=1}$. The class labels are consistent for both support and training sets of the same task, but vary across distinct tasks. Overall the training of MetaNet consists of three main procedures: \textit{acquisition of meta information}, \textit{generation of fast weights} and \textit{optimization of slow weights}, executed collectively by the base and the meta learner. The training of MetaNet is described in Algorithm \ref{alg:metanet}.

To test the model for one-shot SL, we sample another sequence of tasks from a test dataset with unseen classes. Then the model is deployed to classify test examples based on its support set. We assume that we have class labels for the support set during both training and testing. Note that in one-shot learning setup, the support set contains only single example per class and thus it is cheap to obtain.

\subsection{Meta Learner}
\label{meta learner}
The meta learner consists of a dynamic representation learning function $u$ and fast weight generation functions $m$ and $d$. The function $u$ has a representation learning objective and constructs embeddings of inputs in each task space by using task-level fast weights. The weight generation functions $m$ and $d$ are responsible for processing the meta information and generating the example and task-level fast weights.

\begin{figure}[t]
\vskip 0.2in
\begin{center}
\centerline{\includegraphics[width=0.6\columnwidth]{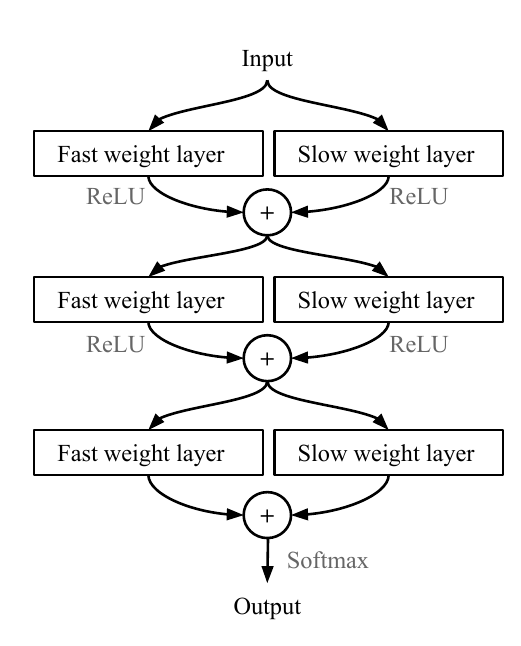}}
\caption{A layer augmented MLP}
\label{layer_aug_fig}
\end{center}
\vskip -0.2in
\end{figure} 

More specifically, the function $m$ learns the mapping from the loss gradient $\lbrace \nabla_{i} \rbrace^N_{i=1}$, derived from the base learner $b$, to fast weights $\lbrace W_i^* \rbrace^N_{i=1}$:
\begin{equation}
\label{fast_w}
W_i^* = m(Z,\nabla_{i})
\end{equation}
where $m$ is a neural network with parameter $Z$.
The fast weights are then stored in a memory $M=\lbrace W_i^* \rbrace^N_{i=1}$. The memory $M$ is indexed with task dependent embeddings $R=\lbrace r'_i \rbrace^N_{i=1}$ of the support examples $\lbrace x'_i \rbrace^N_{i=1}$, obtained by the dynamic representation learning function $u$.

The representation learning function $u$ is a neural net parameterized by slow weights $Q$ and task-level fast weights $Q^*$. It uses the representation loss $loss_{emb}$ to capture a representation learning objective and to obtain the gradients as meta information.
We generate the fast weights $Q^*$ on a per task basis as follows:
\begin{equation}
\label{lossu}
\mathcal{L}_{i} = loss_{emb}(u(Q,x'_i),y'_i)
\end{equation}
\begin{equation}
\nabla_{i}=\nabla_Q \mathcal{L}_i
\end{equation}
\begin{equation}
\label{paramq}
Q^* = d(G, \lbrace \nabla \rbrace^T_{i=1})
\end{equation}
where $d$ denotes a neural net parameterized by $G$, that accepts variable sized input. First, we sample $T$ examples ($T \leq N$) $\lbrace x'_i,y'_i \rbrace^T_{i=1}$ from the support set and obtain the loss gradient as meta information. Then $d$ observes the gradient corresponding to each sampled example and summarizes into the task specific parameters. We use LSTM for $d$ although the order of inputs to $d$ does not matter. Alternatively we can take summation or average of the gradients and use a MLP. However, in our preliminary experiment we observed that the latter results in a poor convergence. 

Once the fast weights are generated, the task dependent input representations $\lbrace r'_i \rbrace^N_{i=1}$ are computed as:
\begin{equation}
\label{embed_u}
r'_i = u(Q,Q^*,x'_i)
\end{equation}
where the parameters $Q$ and $Q^*$ are integrated using the layer augmentation method described in Section \ref{layer_aug}.

The loss, $loss_{emb}$ does not need to be the same as the main task loss $loss_{task}$. However, it should be able to capture a representation learning objective. We use cross-entropy loss when the support set has only a single example per class. When there are more than one examples per class available, contrastive loss \cite{chopra2005learning} is a natural choice for $loss_{emb}$ since both positive and negative samples can be formed. In this case, we randomly draw $T$ number of pairs to observe the gradients and the loss is
\begin{equation}
\mathcal{L}_{i} = loss_{emb}(u(Q,x'_{1,i}),u(Q,x'_{2,i}),l_i)
\end{equation}
where $l_i$ is auxiliary label:
\begin{equation}
    l_i=
    \begin{cases}
      1, & \text{if}\ y'_{1,i}=y'_{2,i} \\
      0, & \text{otherwise}
    \end{cases}
\end{equation}

Once the parameters are stored in the memory $M$ and the memory index $R$ is constructed, the meta learner parameterizes the base learner with the fast weights $W_i^*$. First it embeds the input $x_i$ in the task space by using the dynamic representation learning network (i.e. Equation \ref{embed_u})
and then reads the memory with soft attention:
\begin{equation}
a_i = attention(R,r_i)
\end{equation}
\begin{equation}
W_i^* = norm(a_i)^{\top} M
\end{equation}
where $attention$ calculates similarity between the memory index and the input embedding and we use cosine similarity as $attention$ and $norm$ is a normalization function, for which we use $softmax$. 

\subsection{Base Learner}
\label{base learner}
The base learner, denoted as $b$, is a function or a neural net that estimates the main task objective via a task loss $loss_{task}$. However, unlike standard neural nets, $b$ is parameterized by slow weights $W$ and example-level fast weights $W^*$. The slow weights are updated via a learning algorithm during training whereas the fast weights are generated by the meta learner for every input.

The base learner uses a representation of meta information obtained by using a support set, to provide the meta learner with feedbacks about the new input task.
The meta information is derived from the base learner in form of the loss gradient information:
\begin{equation}
\label{lossb}
\mathcal{L}_{i} = loss_{task}(b(W,x'_i),y'_i)
\end{equation}
\begin{equation}
\nabla_{i}=\nabla_W \mathcal{L}_i
\end{equation}

Here $\mathcal{L}_i$ is the loss for support examples $\lbrace x'_i,y'_i \rbrace^N_{i=1}$. $N$ is the number of support examples in the task set (typically a single instance per class in the one-shot learning setup). $\nabla_{i}$ is the loss gradient with respect to parameters $W$ and is our meta information. Note that the loss function $loss_{task}$ is generic and can take any form, such as a cumulative reward in reinforcement learning. For our one-shot classification setup we use cross-entropy loss. The meta learner takes in the gradient information $\nabla_{i}$ and generates the fast parameters $W^*$ as in Equation \ref{fast_w}.

Assuming that the fast weights $W_{i}^*$ for input $x_i$ are defined, the base learner performs the one-shot classification as:
\begin{equation}
\label{eq_train}
P(\hat y_i|x_i,W,W_{i}^*) = b(W,W_{i}^*,x_i)
\end{equation}
where $\hat y_i$ is predicted output and $\lbrace x_i\rbrace^L_{i=1}$ is an input drawn from the training set $\lbrace x_i,y_i \rbrace^L_{i=1}$ for the current task. Alternatively the base learner can take as input the task specific representations $\lbrace r_i \rbrace^L_{i=1}$ produced by the dynamic representation learning network, effectively reducing the number of MetaNet parameters and leveraging shared representations. In this case, the base learner is forced to operate in the dynamic task space constructed by $u$ instead of building new representations from the raw inputs $\lbrace x_i\rbrace^L_{i=1}$.

During training, given output labels $\lbrace y_i \rbrace^L_{i=1}$, we minimize the cross-entropy loss for one-shot SL.
The training parameters of MetaNet $\theta$ consists of the slow weights $W$ and $Q$ and the meta weights $Z$ and $G$ (i.e. $\theta=\lbrace W,Q,Z,G \rbrace$) and jointly updated via a training algorithm such as backpropagation to minimize the task loss $loss_{task}$ (Equation \ref{eq_train}).

In a similar way, as defined in the Equation \ref{lossu}-\ref{paramq}, we can also parameterize the base learner with task-level fast weights. An ablation experiment on different variation of MetaNet is reported in Section \ref{results}.

\begin{table*}[t]
  \caption{One-shot accuracy on Omniglot previous split
  }
  \label{tab:omni}
  \small
  \centering
  \begin{tabular}{lcccc}
    \toprule
    \bf Model & \bf 5-way & \bf 10-way & \bf 15-way & \bf 20-way             \\
    \midrule
    Pixel kNN \cite{kaiser2017learning} & 41.7 & - & - & 26.7 \\
    Siamese Net \cite{koch2015siamese} & 97.3 & - & - & 88.1 \\
    MANN \cite{santoro2016meta} & 82.8 & - & - & - \\
    Matching Nets \cite{vinyals2016matching} & 98.1 & - & - & 93.8 \\
    Neural Statistician \cite{edwards2016towards} & 98.1 & - & - & 93.2 \\
    Siamese Net with Memory \cite{kaiser2017learning} & 98.4 & - & - & 95.0 \\
    \midrule
     MetaNet- & 98.4 & 98.32 & 96.68 & 96.13 \\
     MetaNet & \bf 98.95 & \bf 98.67 & \bf 97.11 & \bf 97.0 \\
     MetaNet+ & 98.45 & 97.05 & 96.48 & 95.08 \\
    \bottomrule
  \end{tabular}
\end{table*}

\subsection{Layer Augmentation}
\label{layer_aug}
A slow weight layer in the base learner is extended with its corresponding fast weights for rapid generalization. An example of the layer augmentation approach applied to a MLP is shown in Figure \ref{layer_aug_fig}. The input of an augmented layer is first transformed by both slow and fast weights and then passed through a non-linearity (i.e. $ReLU$) resulting in two separate activation vectors. Finally the activation vectors are aggregated by an element-wise vector addition. For the last $softmax$ layer, we first aggregate two transformed inputs and then normalize for classification output.

Intuitively, the fast and slow weights in the layer augmented neural net can be seen as feature detectors operating in two distinct numeric domains. The application of the non-linearity maps them into the same domain, which is $\left[0,\infty \right)$ in the case of $ReLU$ so that the activations can be aggregated and processed further. Our aggregation function here is element-wise sum.

Although it is possible to define the base learner with only fast weights, in our preliminary experiment we found that the integration of both slow and fast weights with the layer augmentation approach is essential in convergence of MetaNet models. A MetaNet model relying on a base leaner with only fast weights were failed to converge and the best performance of this model was reported to be as equal as that of a constant classifier that assigns the same label to every input.

\section{Results}
\label{results}

We carried out one-shot classification experiments on three datasets: Omniglot, Mini-ImageNet and MNIST. The Omniglot dataset consists of images across 1623 classes with only 20 images per class, from 50 different alphabets \cite{lake2015human}. It also comes with a standard split of 30 training and 20 evaluation alphabets. Following \cite{santoro2016meta}, we augmented the training set through 90, 180 and 270 degrees rotations. The images are resized to 28 x 28 pixels for computational efficiency. For the experiment on Mini-ImageNet data, we evaluated on the same class subset provided by \citet{Sachin2017}. MNIST images were used as out-of-domain data. The training details are described in Appendix \ref{training detail}.

\subsection{One-shot Learning Test}
In this section we will report four groups of benchmark experiments: Omniglot previous split, Mini-ImageNet, MNIST as out-of-domain data and Omniglot standard split. 

\subsubsection{Omniglot Previous Split} 
Following the previous setup \citet{vinyals2016matching}, we split the Omniglot classes into 1200 and 423 classes for training and testing. We performed 5, 10, 15 and 20-way one-shot classification and compared our performance against the state-of-the-art results. We also studied three variations of MetaNet as an ablation experiment in order to show how fast parameterization affects the network dynamics. 

In Table \ref{tab:omni}, we compared the performance of our models with all published models (as baselines). The first group of methods are the previously published models. The next group is MetaNet variations. MetaNet is the main architecture described in Section \ref{meta networks}. MetaNet- is a variant without task-level fast weights $Q^*$ in the embedding function $u$ whereas MetaNet+ has additional task-level weights for the base learner in addition to $W^*$. Our MetaNet model improves the previous best results by 0.5\% to 2\% accuracy.
As the number of classes increases (from 5-way to 20-way classification), overall the performance of the one-shot learners decreases. MetaNet's performance drop is relatively small (around 2\%) while the drop for the other models ranges from 3\% to 15\%. As a result, our model shows an absolute improvement of 2\% on 20-way one-shot task. 

Comparing different MetaNet variations, the additional task-level weights in the base learner (MetaNet+) did not seem to help and in fact had a negative effect on performance. MetaNet- however performed surprisingly well but still falls behind the MetaNet model as it lacks the dynamic representation learning function. This performance gap  increases when we test them in out-of-the domain setting (Appendix \ref{mnist_exp}).

\subsubsection{Mini-ImageNet}
The training, dev and testing sets of 64, 16, and 20 ImageNet classes (with 600 examples per class) were provided by \citet{Sachin2017}. By following \citet{Sachin2017}, we sampled 15 examples per class for evaluation. By using the dev set, we set an evaluation checkpoint where only if the model performance exceeds the previous best result on random 400 trials produced from the dev set, we apply the model to another 400 trials randomly produced from the testing set and report the average accuracy.

In Table \ref{tab:mini}, we present the results of the 5-way one-shot evaluation. MetaNet improved the previous result by up to 6\% accuracy and obtained the best result.\footnote{Our code and data will be made available at: \url{ https://bitbucket.org/tsendeemts/metanet}}

\begin{table}[t]
  \caption{One-shot accuracy on Mini-ImageNet test set
  }
  \label{tab:mini}
  \small
  \centering
  \begin{tabular}{lcc}
    \toprule
    \bf Model & \bf 5-way \\
    \midrule
    Fine-tuning \cite{Sachin2017} & 28.86 $\pm$ 0.54 \\
    kNN \cite{Sachin2017} & 41.08 $\pm$ 0.70 \\
    Matching Nets \cite{vinyals2016matching} & 43.56 $\pm$ 0.84\\
    MetaLearner LSTM \cite{Sachin2017} & 43.44 $\pm$ 0.77 \\
    \midrule
     MetaNet & \bf 49.21 $\pm$ 0.96 \\
     \bottomrule
  \end{tabular}
\end{table}

\subsubsection{Omniglot Standard Split}
Omniglot data comes with a standard split of 30 training alphabets with 964 classes and 20 evaluation alphabets with 659 classes. We trained and tested only the standard MetaNet model in this setup. In order to best match the evaluation protocol of \citet{lake2015human}, we form 400 tasks (trials) from the evaluation classes to test the model.

In Table \ref{tab:omni1}, we listed the MetaNet results along with the previous models and human performance. Our MetaNet outperformed the human performance by a slight margin, but underperformed the probabilistic programming approach. However, the performance gap is rather small between these top three baselines. In addition while the probabilistic programming performs slightly better than MetaNet, our model does not
rely on any extra prior knowledge about how characters and strokes are composed. Comparing the results on two Omniglot splits in Tables \ref{tab:omni} and \ref{tab:omni1}, MetaNet showed decreasing performances on the standard split. The later setup seems to be slightly difficult as the number of classes in the training set is less (1200 vs 964) and test classes are bigger (423 vs 659).

\begin{table*}[t]
  \caption{One-shot accuracy on Omniglot standard split
  }
  \label{tab:omni1}
  \small
  \centering
  \begin{tabular}{lcccc}
    \toprule
    \bf Model & \bf 5-way & \bf 10-way & \bf 15-way & \bf 20-way             \\
    \midrule
    Human performance \cite{lake2015human} & - & - & - & 95.5 \\
    \midrule
    Pixel kNN \cite{lake2013one} & - & - & - & 21.7 \\
    Affine model \cite{lake2013one} & - & - & - & 81.8 \\
    Deep Boltzmann Machines \cite{lake2013one} & - & - & - & 62.0 \\
    
    Hierarchial Bayesian Program Learning \cite{lake2015human} & - & - & - & \bf 96.7 \\
    Siamese Net \cite{koch2015siamese} & - & - & - & 92.0 \\
    \midrule
     MetaNet & 98.45 & 97.32 & 96.4 & 95.92 \\
     \bottomrule
  \end{tabular}
\end{table*}

\subsection{Generalization Test}
We conducted a set of experiments to test the generalization of MetaNet from multiple aspects. The first experiment tests whether a MetaNet model trained on an N-way one-shot task could generalize to another K-way task (where $N \neq K$) without actually training on the second task. The second experiment is to test if a meta learner trained for rapid parameterization of a base learner $b_{train}$ could parameterize another base learner $b_{eval}$ during evaluation. The last experimental setup examines whether MetaNet supports meta-level continual learning.

\begin{table}[b]
  \caption{Accuracy of MetaNet trained on N-way and tested on K-way one-shot tasks}
  \label{tab:nkway}
  \small
  \centering
  \begin{tabular}{lcccc}
    \toprule
    {} & \multicolumn{4}{c}{\bf Test}   \\
  \cmidrule{2-3} \cmidrule{4-5}
    \bf Train & 5-way & 10-way & 15-way & 20-way  \\
    \midrule
    5-way & 98.95 & 96.4 & 93.6 & 93.07 \\
    10-way & 99.25 & 96.87 & 96.95 & 96.21  \\
    15-way & 99.35 & 98.17 & 97.11 & 96.36 \\
    20-way & 99.55 & 98.87 & 97.41 & 97.0 \\
    \bottomrule
  \end{tabular}
\end{table}

\subsubsection{N-way Training and K-way Testing}
In this experiment, MetaNet is trained on N-way one-shot classification task and then tested on K-way one-shot tasks. The number of training and test classes are varied (i.e. $N \neq K$). To handle this, we inserted a $softmax$ layer into the base learner during evaluation and then augmented it with the fast weights generated by the meta learner. If the meta learner is generic enough, it should be able to parameterize the new $softmax$ layer on the fly. The new layer weights remained fixed since no parameter update was performed for this layer. The K-way test tasks were formed from the 423 unseen classes in the test set.

The MetaNet models were trained on one of 5, 10, 15 and 20-way one-shot tasks and evaluated on the rest. In Table \ref{tab:nkway}, we summarized the results. As a comparison we also included some results from Table \ref{tab:omni}, which reports accuracy of N-way train and test setting. The MetaNet model trained on 5-way tasks obtained 93.07\% of 20-way test accuracy which is still a closer match to Matching Network and higher than Siamese Net trained 20-way tasks. An interesting finding is that when $N$ is smaller than $K$, i.e. the model is trained on easier tasks than test ones, we observe a decreasing performance. Conversely the models trained on harder tasks (i.e. $N>K$) achieved increasing performances when tested on the easier tasks and the performance is even higher than the ones that were applied to the tasks with the same level difficulty (i.e. $N=K$). For example, the model skilled on 20-way classification improved the 5-way one-shot baseline by 0.6\% showing a ceiling performance in this setting. We also conducted a preliminary experiment on more extreme test-time classification. MetaNet trained on 10-way task achieved around 65\% on 100-way one-shot classification task.

This flexibility in MetaNet is crucial because one-shot learning usually involves an online concept identification scenario. Furthermore we can empirically obtain a performance lower or upper bound. Particularly the test performance obtained on the tasks with the same level difficulty that the model was skilled on can be used as a performance lower or an upper bound depending on a scenario under which the model will be deployed in the future. For example, for the MetaNet model that will deployed under the $N>K$ scenario, we can obtain the performance lower bound by testing it on the $N=K$ tasks.

\subsubsection{Rapid Parameterization of Fixed Weight Base Learner}
We replaced the entire base learner with a new CNN during evaluation. The slow weights of this network remained fixed. The fast weights are generated by the meta learner that is trained to parameterize the old base learner and used to augmented the fixed slow weights.

\begin{figure}[t]
\vskip -0.1in
\begin{center}
\centerline{\includegraphics[width=\columnwidth]{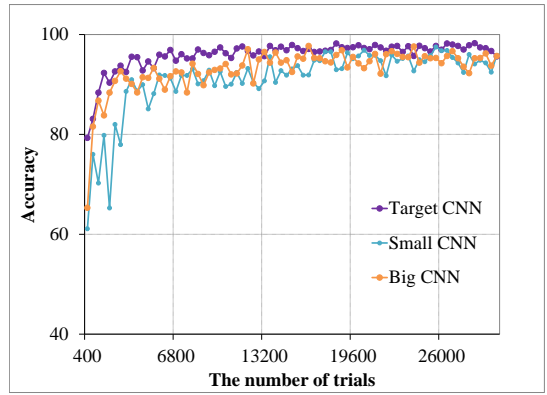}}
\vskip -0.2in
\caption{Comparison of the test performances of the base learners on Omniglot 5-way classification.}
\label{meta_acc}
\end{center}
\vskip -0.3in
\end{figure}

We tested a small and a large CNN for the base learner. The small CNN has 32 filters and the large CNN has 128 filters. 
In Figure \ref{meta_acc}, the test performances of these CNNs are compared. The base learner (target CNN) optimized along within the model performed better than the fixed weight CNNs. The performance difference between these models is large in earlier training iterations. However, as the meta learner sees more one-shot learning trials, the test accuracies of the base learners converge. This results show that MetaNet effectively learns to parameterize a neural net with fixed weights.

\subsubsection{Meta-Level Continual Learning}
MetaNet operates in two spaces: input problem space and meta (gradient) space. If the meta space is problem independent, MetaNet should support meta-level continual learning or life-long learning. This experiment tests this in the case of the loss gradient. 

Following the previous work on catastrophic forgetting in neural networks \cite{srivastava2013compete,goodfellow2013empirical,kirkpatrick2016overcoming}, we formulated two problems in a sequential manner. We first trained and tested the model on the Omniglot sets and then we switched and continued training on the MNIST data. After training on a number of MNIST one-shot tasks, we re-evaluated the model on the same Omniglot test set and compare performance. A decrease in performance here indicates that the meta weights $Z$ and $G$ of the neural nets $m$ and $d$ are prone to catastrophic forgetting and the model therefore does not support continual learning. On the other hand, an increase in performance indicates that MetaNet supports reverse transfer learning and continual learning. 

\begin{figure}[t]
\vskip -0.1in
\begin{center}
\centerline{\includegraphics[width=\columnwidth]{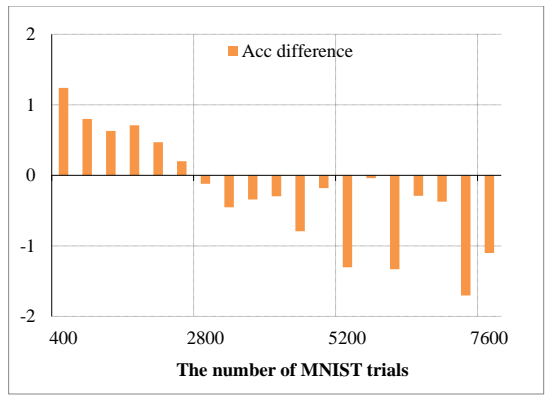}}
\vskip -0.2in
\caption{The difference between the two Omniglot test accuracies obtained before and after training on MNIST task.}
\label{meta_con}
\end{center}
\vskip -0.4in
\end{figure} 

We allocated separate parameters for the weights $W$ and $Q$ when we switched the problems so the only meta weights were updated. We used two three-layer MLPs with 64 hidden units as the embedding function and the base learner. The MNIST image and classes were augmented by randomly permuting the pixels. We created 50 different random shuffles and thus the training set for the second one-shot problem consisted of 500 classes. We conducted multiple runs and increased the MNIST training trials by multiples of 400 (i.e. 400, 800, 1200...) in each run giving more time for MetaNet to adapt its meta weights on the second problem so that it may forget the knowledge about Omniglot. Each run was repeated five times and we report the average statistics. For every run, the network and the optimizer were reinitialized and the training started from scratch.

In Figure \ref{meta_con}, we plotted the accuracy difference between two Omniglot test performances obtained before and after training on the MNIST task. The performance improvement (y-axis) after training on the MNIST tasks ranges from -1.7\% to 1.24\% depending on the training time (x-axis). The positive values indicate that the training on the second problem automatically improves the performance of the earlier task exhibiting the reverse transfer property. Therefore, we can conclude that MetaNet successfully performs reverse transfer. At the same time, it is skilled on MNIST one-shot classification. The MNIST training accuracy reaches over 72\% after 2400 MNIST trials. However, reverse transfer happens only up to a certain point in MNIST training (2400 trials). After that, the meta weights start to forget the Omniglot information. As a result from 2800 trials onwards, the Omniglot test accuracy drops. Nevertheless even after 7600 MNIST trials, at which point the MNIST training accuracy reached over 90\%,  the Omniglot performance drop was only 1.7\%.

\section{Discussion and Future Work}
One-shot learning in combination with a meta learning framework can be a useful approach to address certain neural network drawbacks related to rapid generalization with small data and continual learning. We present a novel meta learning method, MetaNet, that performs a generic knowledge acquisition in a meta space and shifts the parameters and inductive biases of underlying neural networks via fast parameterization for the rapid generalization.

Under the MetaNet framework, an important consideration is the type of higher order meta information that can be extracted as a feedback from the model when operating on a new task. One desirable property here is that the meta information should be generic and problem independent. It should also be expressive enough to explain the model setting in the current task space. We explored the use of loss gradients as meta information in this work. As shown in the results, using the gradients as meta information seems to be a promising direction. MetaNet obtains state-of-the art results on several one-shot SL benchmarks and leads to a very flexible AI model. For instance, in MetaNet we can alternate between different $softmax$ layers on the fly during test. It supports continual learning up to a certain point. We observed that neural nets with fixed slow weights can perform well for new task inputs when augmented with the fast weights. When the slow weights are updated during training, it learns domain biases resulting in even better performance on identification of new concepts within the same domain. However, one could expect a higher performance from the fixed weight network when aiming for one-shot generalization across distant domains.

An interesting future direction would be in exploring a new type of meta information that is more robust and expressive, and in developing synaptic weights that are capable of maintaining such higher order information. One could take inspiration from the meta learning process in the brain and ask whether the brain operates on some kind of higher order information to generalize across tasks and acquire new skills. 

The rapid parameterization approach presented here has been shown to be an effective alternative to the direct optimization methods that learn to update network parameters for one-shot generalization. However, a problem this approach poses is the integration of slow and fast weights. As a solution to this, we presented a simple layer augmentation method. Although the layer augmentation worked reasonably well, this method becomes difficult when a neural net has many types of parameters operating in multiple different time-scales. For example, a single base learner equipped with three types of weights (slow, example-specific, and task-level weights) integrated under the layer augmentation paradigm could not perform as well as a simpler one. Therefore, a potential extension would be to train MetaNet so it can discover its own augmentation schema for efficiency.

MetaNet can readily be applied to parameterize policies in reinforcement learning and imitation learning, leading to an agent with one-shot and meta learning capabilities. MetaNet based on recurrent networks as underlying learners could lead to useful applications in sequence modeling and language understanding tasks.

\section*{Acknowledgements} 
We would like to thank the anonymous reviewers and our colleagues, Jesse Lingeman, Abhyuday Jagannatha and John Lalor for their insightful comments and suggestions on improving the manuscript.
This work was supported in part by the grant HL125089 from the National Institutes of Health and by the grant 1I01HX001457-01 supported by the Health Services Research \& Development of the US Department of Veterans Affairs Investigator Initiated Research. Any opinions, findings and conclusions or recommendations expressed in this material are those of the authors and do not necessarily 
reflect those of the sponsor.


\bibliography{example_paper}
\bibliographystyle{icml2017}

\clearpage

\appendix

\section{Training Details}
\label{training detail}
To train and test MetaNet on one-shot learning, we adapted the training procedure introduced by \citet{vinyals2016matching}. First, we split the data into training and test sets consisting of two disjoint classes. We then formulate a series of tasks (trials) from the training set. Each task has a support set of $N$ classes with one image per, resulting in an N-way one-shot classification problem. In addition to the support set, we also include $L$ number of labeled examples in each task set to update the parameters $\theta$ during training. For testing, we follow the same procedure to form a set of test tasks from the disjoint classes. However, now MetaNet assigns class labels to $L$ examples based only on the labeled support set of each test task.

For the one-shot benchmarks on the Omniglot dataset, we used a CNN with 64 filters as the base learner $b$. This CNN has 5 convolutional layers, each of which is a 3 x 3 convolution with 64 filters, followed by a $ReLU$ non-linearity, a 2 x 2 max-pooling layer, a fully connected (FC) layer, and a softmax layer. Another CNN with the same architecture is used to define the dynamic representation learning function $u$, from which we take the output of the FC layer as the task dependent representation $r$. We trained a similar CNNs architecture with 32 filters for the experiment on Mini-ImageNet. However for computational efficiency as well as to demonstrate the flexibility of MetaNet, the last three layers of these CNN models were augmented by fast weights. For the networks $d$ and $m$, we used a single-layer LSTM with 20 hidden units and a three-layer MLP with 20 hidden units and $ReLU$ non-linearity. As in \citet{andrychowicz2016learning}, the parameters $G$ and $Z$ of $d$ and $m$ are shared across the coordinates of the gradients $\nabla$ and the gradients are normalized using the same preprocessing rule (with $p=7$). The MetaNet parameters $\theta$ are optimized with ADAM. The initial learning rate was set to $10^{-3}$. The model parameters $\theta$ were randomly initialized from the uniform distribution over [-0.1, 0.1).

\section{MNIST as Out-Of-Domain Data}
\label{mnist_exp}
We treated MNIST images as a separate domain data. Particularly a model is trained on the Omniglot training set and evaluated on the MNIST test set in 10-way one-shot learning setup. We hypothesize that models with a high dynamic should perform well on this task.

\begin{figure}[h]
\vskip -0.1in
\begin{center}
\centerline{\includegraphics[width=\columnwidth]{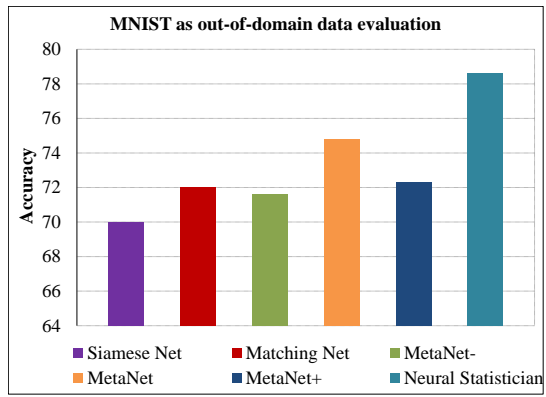}}
\vskip -0.2in
\caption{MNIST 10-way one-shot classification results.}
\label{mnist_eval}
\end{center}
\vskip -0.2in
\end{figure}

In Figure \ref{mnist_eval}, we plotted the results of this experiment. MetaNet- achieved 71.6\% accuracy which was 0.6\% and 3.2\% lower than the other variants with fast weights. This is not surprising since MetaNet without dynamic representation learning function lacks an ability to adapt its parameters to MNIST image representations. The standard MetaNet model achieved 74.8\% and MetaNet+ obtained 72.3\%. Matching Net \cite{vinyals2016matching} reported 72.0\% accuracy in this setup. Again we did not observe improvement with MetaNet+ model here. 
The best result was recently reported by using a generative model, Neural Statistician, that extends variational autoencoder to summarize input set \citep{edwards2016towards}.

\end{document}